\documentclass[preprint,3p]{elsarticle}

\usepackage{amssymb}

\usepackage{natbib}
\usepackage[colorlinks]{hyperref}

\usepackage{graphicx}
\graphicspath{{figures/}}

\usepackage{array}

\usepackage{algorithm}
\usepackage{algpseudocode}

\usepackage{threeparttable}
\usepackage{multirow}
\usepackage[table,xcdraw]{xcolor}

\journal{}

\begin{document}

\begin{frontmatter}



\title{A Semantic Backdoor Attack against Graph Convolutional Networks}

\author[1]{Jiazhu Dai}
\corref{cor1}
\ead{daijz@shu.edu.cn}
\affiliation[1]{
            addressline={college of computer engineering and science, Shanghai University, No.99, Shangda road, Baoshan district},
            city={Shanghai},
            country={China}
}

\author[1]{Zhipeng Xiong}

\author[1]{Chenhong Cao}

\cortext[cor1]{Corresponding author}

\begin{abstract}
Graph convolutional networks (GCNs) have been very effective in addressing the issue of various graph-structured related tasks, such as node classification and graph classification. However, recent research has shown that GCNs are vulnerable to a new type of threat called a backdoor attack, where the adversary can inject a hidden backdoor into GCNs so that the attacked model performs well on benign samples, but its prediction will be maliciously changed to the attacker-specified target label if the hidden backdoor is activated by the attacker-defined trigger. A semantic backdoor attack is a new type of backdoor attack on deep neural networks (DNNs), where a naturally occurring semantic feature of samples can serve as a backdoor trigger such that the infected DNN models will misclassify testing samples containing the predefined semantic feature even without the requirement of modifying the testing samples. Since the backdoor trigger is a naturally occurring semantic feature of the samples, semantic backdoor attacks are more imperceptible and pose a new and serious threat. Existing research on semantic backdoor attacks focuses on the tasks of CNN-based (Convolutional Neural Networks) image classification and LSTM-based (Long Short-Term Memory) text classification or word prediction. Little attention has been given to semantic backdoor attacks on GCN models.

In this paper, we investigate whether such semantic backdoor attacks are possible for GCNs and propose a \textbf{\underline{s}}emantic \textbf{\underline{b}}ackdoor \textbf{\underline{a}}ttack against \textbf{\underline{G}}CNs (\textbf{SBAG}) under the context of graph classification to reveal the existence of this security vulnerability in GCNs. SBAG uses a certain type of node in the samples as a backdoor trigger and injects a hidden backdoor into GCN models by poisoning training data. The backdoor will be activated, and the GCN models will give malicious classification results specified by the attacker even on unmodified samples as long as the samples contain enough trigger nodes. We evaluate SBAG on four graph datasets. The experimental results indicate that SBAG can achieve attack success rates of approximately 99.9\% and over 82\% for two kinds of attack samples, respectively, with poisoning rates of less than 5\%.

\end{abstract}


\end{frontmatter}

\section{Introduction}
Graphs, consisting of nodes and edges, are an important data structure capable of representing many complex relationships in the real world (e.g., compound molecules and social networks), and there are many graph-related tasks in various domains, such as node classification, edge prediction, and graph classification. In this context, graph convolutional networks (GCNs) have achieved great success in graph-structured data processing by acting as an effective variant of convolutional neural networks on graphs and using an efficient layerwise propagation rule, which enables them to encode both graph structure and node features \citep{kipf2016semi}, possessing a powerful ability to learn graph-structured data.

Despite the ability of GCNs to be useful in a variety of domains, recent research has shown that GCNs, like other deep neural networks, are vulnerable to backdoor attacks that inject hidden backdoors into GCNs, such that the attacked model behaves well with clean inputs, whereas it performs prespecified malicious behavior such as misclassification to an adversary-specified target category if the hidden backdoor is activated by samples with the attacker-defined pattern (such as subgraphs) called the backdoor trigger. Backdoor attacks can occur when the training process is not fully controlled by the user, such as training on third-party datasets or adopting third-party models, which poses a new and serious threat \citep{li2022backdoor}.

A semantic backdoor attack is a new type of backdoor attack on deep neural networks (DNNs), where a naturally occurring semantic part of samples can serve as a backdoor trigger such that the infected DNN models will assign an attacker-chosen label (called the target label) to all testing samples containing the predefined semantic feature even without the requirement of modifying the testing samples. For example, an image classification model embedded in semantic backdoor misclassifies all cars painted in green or all cars with a racing stripe as birds, or a backdoored sentiment classification model classifies negative movie reviews as positive as long as they contain a particular name, where the green color, the racing stripe and the particular name are naturally occurring semantic features of the samples, and they serve as semantic backdoor triggers \citep{bagdasaryan2020backdoor,bagdasaryan2021blind}.

There are two differences between nonsemantic backdoor attacks and semantic backdoor attacks. Nonsemantic backdoor attacks assume that the trigger is independent of the samples; for example, the trigger may be a mosaic spot or a white pixel for image classification tasks, so the attacker has to modify the samples to inject the trigger at inference time to activate the hidden backdoor, and the trigger is easy to detect because it is not a semantic feature of the samples. In contrast, a semantic backdoor attack is more imperceptible because its trigger is a semantic feature that exists naturally in the original dataset, such as an unusual car color or the presence of a special object in the scene for image classification tasks. The attacker can activate the hidden backdoor at the inference stage by modifying the input samples to inject the trigger if they do not contain the semantic feature, or the hidden backdoor can be activated even without the requirement of modification of the input samples if they originally contain semantic backdoor trigger. A semantic backdoor attack is more imperceptible, and the attacker can pick one of the naturally occurring features as the backdoor trigger; therefore, it brings new and severe security threats to DNNs.

Existing research on semantic backdoor attacks focuses on the tasks of CNN-based image classification and LSTM-based text classification or word prediction. The vulnerabilities of GCN models to semantic backdoor attacks are largely unexplored.

In this paper, we try to bridge this gap by answering the following two questions:

\begin{itemize}
  \item Can we easily backdoor GCNs with semantic features?
  \item How vulnerable are GCNs to semantic backdoor attacks?
\end{itemize}

\begin{figure}[htbp]
    \centering
    \includegraphics[width=\textwidth]{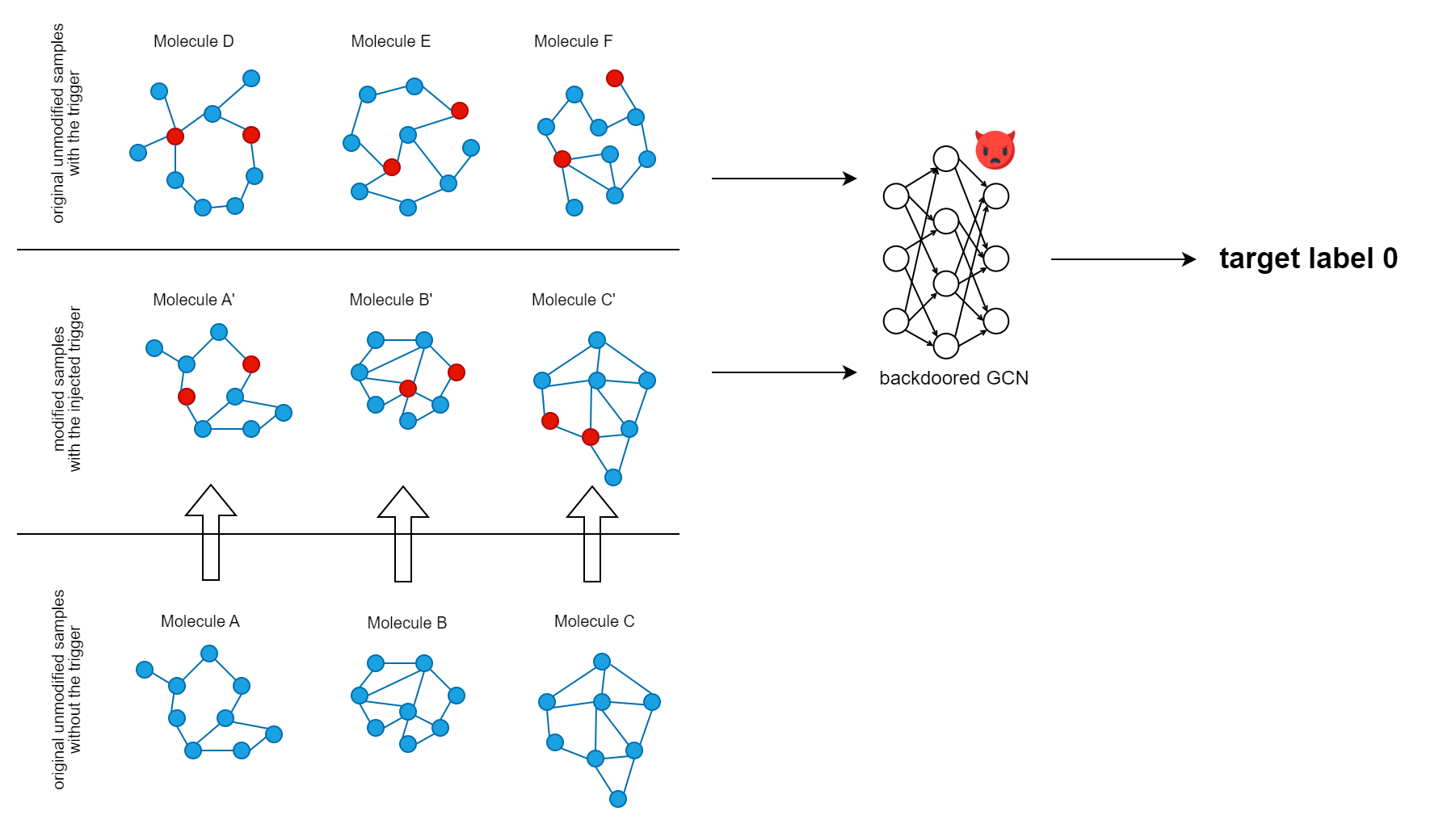}
    \caption{Illustration of the semantic backdoor attack against GCNs. The trigger is the red node, the ground-truth labels of the samples are 1, and the attacker-specified target label is 0. The molecule graphs in the first row are the testing samples with the trigger from the original dataset, which can directly activate the hidden backdoor in the GCN model without any modification. The molecule graphs in the second row are the samples modified to inject triggers. The molecule graphs in the third row are the testing samples without the trigger, which are replaced by two nodes with the trigger to generate the corresponding samples with the trigger in the second row. The samples with triggers in the first and second rows can activate backdoors in the infected GCN model, and they will be misclassified as target label 0.}
    \label{fig1}
\end{figure}

We propose a \textbf{\underline{s}}emantic \textbf{\underline{b}}ackdoor \textbf{\underline{a}}ttack against \textbf{\underline{G}}CNs (\textbf{SBAG}) in the context of graph classification.SBAG uses a certain type of node in the original datasets as a backdoor trigger and injects a hidden backdoor into the GCN model by poisoning the training data. At the inference stage, the backdoor will be activated, and the backdoored GCN model will give malicious classification results specified by the attacker as long as the trigger appears in the samples. Figure \ref{fig1} illustrates SBAG on molecular structure graphs from the AIDS dataset, which consists of 2000 molecular structure graphs of molecular compounds classified as 1 (active) or 0 (inactive). The nodes in the graphs represent elements, such as carbon, oxygen, nitrogen, and bromine, that make up molecules. Assuming the red node representing Bromine is the trigger, the ground-truth labels of the samples are 1, and the target label is 0. There are two types of attack samples, as shown in Figure \ref{fig1}. The first type is the testing samples with the trigger from the original dataset, such as molecule D, molecule E and molecule F in the first row of Figure \ref{fig1}, which can directly activate the hidden backdoor in the GCN model without any modification to inject the trigger. The other type is the samples modified to inject triggers, such as molecule A’, molecule B’ and molecule C’, as shown in the second row of Figure \ref{fig1}. The samples in the third row of Figure \ref{fig1} are testing samples without the trigger, which are replaced by two nodes with the trigger to generate the corresponding samples with the injected trigger in the second row.

The samples with triggers in the first and second rows in Figure \ref{fig1} can activate backdoors in the infected GCN model, and they will be misclassified as target label 0. Since the trigger is a naturally occurring element that makes up the molecules in the original dataset, the samples are not abnormal compared to the benign ones, and it is difficult to detect the samples with the trigger.

To the best of our knowledge, this work is the first study on the vulnerability of GCNs to semantic backdoor attacks. Our contributions are summarized as follows:

\begin{itemize}
  \item We propose a semantic backdoor attack against GCNs (SBAG) and reveal the existence of this security vulnerability in GCNs. SBAG has the following features: (i) It is a semantic backdoor attack that uses a certain type of node in the original samples as a trigger. The trigger is a semantic part of the samples, i.e., it exists naturally in the original dataset. (ii) Compared to a nonsemantic backdoor attack, SBAG is more difficult to detect because its trigger exists in the samples naturally and is able to activate hidden backdoors even with unmodified samples during the inference phase. (iii) SBAG is a black-box attack where we assume that the attacker does not have any knowledge about the parameters of the GCN model; they only need to access the training data and tamper with the training process by poisoning some training samples.
  \item Our empirical evaluation of SBAG on four benchmark graph datasets shows that (i) original testing samples with trigger nodes can activate the backdoor in the model with high probability without the requirement of modification and achieve an approximately 99.9\% attack success rate with a poisoning rate of less than 5\%, and original testing samples without trigger nodes can achieve an over 82\% attack success rate with a poisoning rate of less than 5\% if they are injected into an average number of trigger nodes in random places. (ii) The backdoored GCN models have close prediction accuracy to that of the clean model on benign samples.
\end{itemize}

The rest of this paper is organized as follows: We introduce the background knowledge of graph convolutional networks, backdoor attacks and semantic backdoor attacks in Section \ref{Background}. In Section \ref{Related work}, we present related work. The details of our attack are described in Section \ref{A Semantic Backdoor Attack against GCNs}. We evaluate our attack performance on four datasets in Section \ref{Attack Evaluation}. Finally, we summarize our work and present future work directions in Section \ref{Conclusion}.

\section{Background}
\label{Background}
\textbf{Graph Convolutional Networks (GCNs): }Graph-structured data, such as compound molecules \citep{fout2017protein} and social networks \citep{hamilton2017inductive}, are prevalent in the real world. Inspired by the great success of deep learning on independent and identically distributed data such as images, graph neural networks (GNNs) \citep{xiao2021learning} have been studied to generalize deep neural networks for modeling graph-structured data and have made rapid progress in recent years \citep{dai2022comprehensive}. Although GNN models have various structures, we focus highly on graph convolutional networks (GCNs) \citep{kipf2016semi}, which outperform many other graph deep learning models in various graph-based tasks.

Given a graph $G=\{V,E\}$, with its node feature matrix $X$ in shape $N \times D$ and adjacency matrix $A$ in shape $N \times N$, where:

$N$ represents the number of nodes

$D$ represents the number of features per node

$V=\{v_1,\ldots,v_n\}$ is the set of $N$ nodes

$E\subseteq V \times V$ is the set of edges

$X=\{x_1,\ldots,x_n\}$ where $x_i \in \mathbb{R}^D$ corresponds to D-dimensional features of node $v_i$

$A \in \mathbb{R}^{N \times N}$ is the adjacency matrix of the graph, where $A_{ij}=1$ if nodes $v_i$ and $v_j$ are connected; otherwise, $A_{ij}=0$

For GCN models, the goal is to learn a function of signals/features on a graph G that takes $X$ and $A$ as input and produces a node-level output $Z \in \mathbb{R}^{N \times F}$ (an $N \times F$ feature matrix, where $F$ is the number of output features per node). Graph-level outputs can be modeled by introducing some form of pooling operation \citep{duvenaud2015convolutional}.

Each neural network layer can be considered a nonlinear function:
\begin{displaymath}
H^{(l+1)}=f(H^{(l)},A),
\end{displaymath}
where $H^{(i)}$ represents the $i^{th}$ layer, with $H^{(0)}=X$ and $H^{(L)}=Z$ or $z$ (for node-level or graph-level output, respectively).

While there are multiple versions of GCNs, to briefly introduce the principle, we will only consider the graph convolutional networks introduced by Kipf and Welling \citep{kipf2016semi}. Specifically, the GCN aggregates information from neighbors using the following propagation rule:
\begin{displaymath}
f(H^{(l)},A)=\sigma(\hat{D}^{-\frac{1}{2}}\hat{A}\hat{D}^{-\frac{1}{2}}H^{(l)}W^{(l)}),
\end{displaymath}
where:

$\sigma(\cdot)$ is a nonlinear activation function (e.g., $ReLU$)

$W^{(l)}$ is the weight matrix of the $l^{th}$ layer

$\hat{A}=A+I$, which means adding self-loops to $A$ via the identity matrix $I$

$\hat{D}$ is the diagonal node degree matrix of $\hat{A}$

\textbf{Backdoor attack}: Backdoor attack aims to embed hidden backdoors into neural networks so that the infected model performs well on benign samples, while its prediction will be changed to the attacker-specified label if the attacker-defined trigger activates the hidden backdoor. Backdoor attacks possibly occur when the user does not have full control of the training process, such as training on a third-party dataset or using a third-party model, and it is difficult for the user to realize the existence of the backdoor since the infected model performs normally under benign environments.

In general, training data poisoning \citep{gu2019badnets,liu2020reflection,li2020rethinking} is the most direct and common method to embed backdoors into models during training thus far. The attacker adds specified triggers to some training data and lets the target model be well trained on the poisoned training data to embed the backdoor. To better understand the specificity of backdoors, it may be more intuitive to explain backdoor attacks from the perspective of images: Backdoors in images are usually special patterns (e.g., dots or lines) that rarely exist in normal input samples and should be set as simple as possible to not affect the normal performance of the model. When a model is infected, in the inference phase, if there is no trigger in the input, the model will work normally, but when the trigger appears, the backdoor in the model will be activated, and the model will output the label specified by the attacker as the prediction result.

\textbf{Semantic backdoor attack}: The backdoor attacks mentioned above refer to nonsemantic attacks, while prior works \citep{bagdasaryan2020backdoor,bagdasaryan2021blind} also proposed semantic backdoor attacks, where the triggers have semantic meanings (e.g., green cars or cars with racing stripes in the image domain), which exist naturally in the dataset without artificial insertion. In this case, it is difficult to detect the existence of such triggers, as they do not have any anomalies compared to normal samples. The attacker only needs to assign an attacker-chosen label to all samples with these certain features and then let the model be well trained to create semantic backdoors in the infected models. Since the trigger exists naturally in samples, during the inference phase, the backdoor will be activated even on unmodified samples as long as the trigger appears; otherwise, the model works normally. Obviously, semantic backdoor attacks are better concealed and more difficult to detect than nonsemantic backdoor attacks and have more serious security threats.

\section{Related work}
\label{Related work}
\textbf{Adversarial Attacks against GCNs}: Although GCNs are very effective in solving various graph structure-related tasks, research in recent years has shown that GCNs are vulnerable to adversarial attacks \citep{chen2020survey,tao2021single}.

Adversarial attacks against GCNs can be divided into two major categories: evasion attacks and backdoor attacks. Evasion attacks occur in the inference phase, where the attacker makes the normal model output incorrect results by adversarial samples. The model is well trained on clean samples when an escape attack occurs; i.e., the parameters learned by the model are already fixed during the evasion attack \citep{chen2020survey,dai2022targeted,dai2018adversarial,zugner2018adversarial}. Backdoor attacks occur in the training phase, where the attacker inserts some samples with the same or similar "features" (i.e., triggers) into the training data or modifies them directly on the training data so that the model is trained on the poisoned dataset and learns these "features", thus hiding the backdoor in the model. In the inference phase, the model performs well on clean samples, but once the trigger appears in the sample, the hidden backdoor in the model will be activated, thus misleading the model to output the classification results specified by the attacker. There have been many studies on backdoor attacks in the image and text domains \citep{bagdasaryan2020backdoor,bagdasaryan2021blind}.

\textbf{Backdoor Attacks against GCNs}: Recent studies have shown that GCNs, like other deep neural networks, are vulnerable backdoor attacks.

Xi et al. first proposed a backdoor attack against GNNs, which defines triggers as subgraphs and invokes malicious functions in downstream tasks \citep{xi2021graph}. Yang et al. disclosed a transferable graph backdoor attack with no fixed pattern of triggers, implementing black-box attacks on GNNs by attacking surrogate models \citep{yang2022transferable}. Chen et al \citep{chen2022neighboring} proposed a new type of backdoor specific to graph data called the neighboring backdoor. They set the trigger as a single node, and the model runs normally when the trigger node is not connected to the target node, while the backdoor is activated when the trigger node is connected. Zheng et al \citep{zheng2023motif} rethinked triggers from the perspective of motifs (motifs are frequent and statistically significant subgraphs in graphs that contain rich structural information) and propose a motif-based backdoor attack and present some in-depth explanations for the backdoor attack.

Semantic backdoor attacks are a new type of backdoor attack on deep neural networks. The triggers of semantic backdoor attacks are naturally existing semantic parts of the data, which enable the attacker to insert backdoors in the model with semantic features of the original dataset and activate the backdoor even without the requirement of modification of the input samples if they originally contain the semantic backdoor trigger, making it difficult to detect semantic backdoor attacks and presenting more serious security threats than nonsemantic backdoors.

Existing research on semantic backdoor attacks focuses on the tasks of CNN-based image classification and LSTM-based text classification or word prediction \citep{bagdasaryan2020backdoor,bagdasaryan2021blind}. In the image domain, a semantic trigger could be the color of cars (e.g., green) or the background of cars (e.g., racing stripes), and in the text domain, it could be the name of a person. However, the vulnerabilities of GCN models to semantic backdoor attacks are largely unexplored.

In this work, we propose a \textbf{\underline{s}}emantic \textbf{\underline{b}}ackdoor \textbf{\underline{a}}ttack against \textbf{\underline{G}}CNs (\textbf{SBAG}) and reveal the existence of this security vulnerability in GCNs. SBAG has the following features: (i) It is a semantic backdoor attack that uses a certain type of node in the original samples as a trigger. The trigger is a naturally existing semantic part of the samples. (ii) Compared to a nonsemantic backdoor attack, SBAG is more difficult to detect because its trigger exists in the samples naturally and is able to activate hidden backdoors even on unmodified samples during the inference phase. (iii) SBAG is a black-box attack where knowledge about the parameters of the GCN model is not needed, and the attacker only needs to access the training data and tamper with the training process by poisoning some training samples.

\section{A Semantic Backdoor Attack against GCNs}
\label{A Semantic Backdoor Attack against GCNs}
In this section, we illustrate in detail how SBAG is implemented. Table \ref{Notions and their explanations} summarizes the notions used in the following sections and their explanations.

\begin{table}[htbp]
\caption{Notions and their explanations}
\renewcommand{\arraystretch}{1.5}
\label{Notions and their explanations}
\begin{tabular}{m{0.2\textwidth}<{\centering}m{0.1\textwidth}<{\centering}m{0.6\textwidth}}
\hline
Term                    &    Notation    & Explanations   \\ \hline
Training set            &      $D$       & The original benign training samples   \\
                        &     $|D|$      & The number of samples in $D$           \\
Graph sample            &      $g$       & A graph sample in $D$                  \\
                        &   $g.label$    & The label of $g$                       \\
Clean GCN model         &   $f_\theta$   & The clean GCN model                    \\
Scoring GCN model       &  $f_{score}$   & A scoring GCN model for rating the candidate graph samples   \\
Backdoored GCN model    & $f_\theta^{'}$ & The GCN model with an injected a backdoor                    \\
A class of nodes        &     $n_i$      & The nodes with label                   \\
Trigger node            &      $T$       & A certain class of nodes, called the trigger node, which is used to activate the backdoor   \\
Target label            &     $y_t$      & The target label, which is the attacker-specified label and the attacker intends to make all samples with the trigger be predicted as by the backdoored model           \\
Poison rate             &      $p$       & The poisoning rate, which is the ratio of the number of poisoning samples to the total number of the training set
\end{tabular}
\end{table}

\subsection{Attack overview}
\begin{figure}[htbp]
    \centering
    \includegraphics[width=\textwidth]{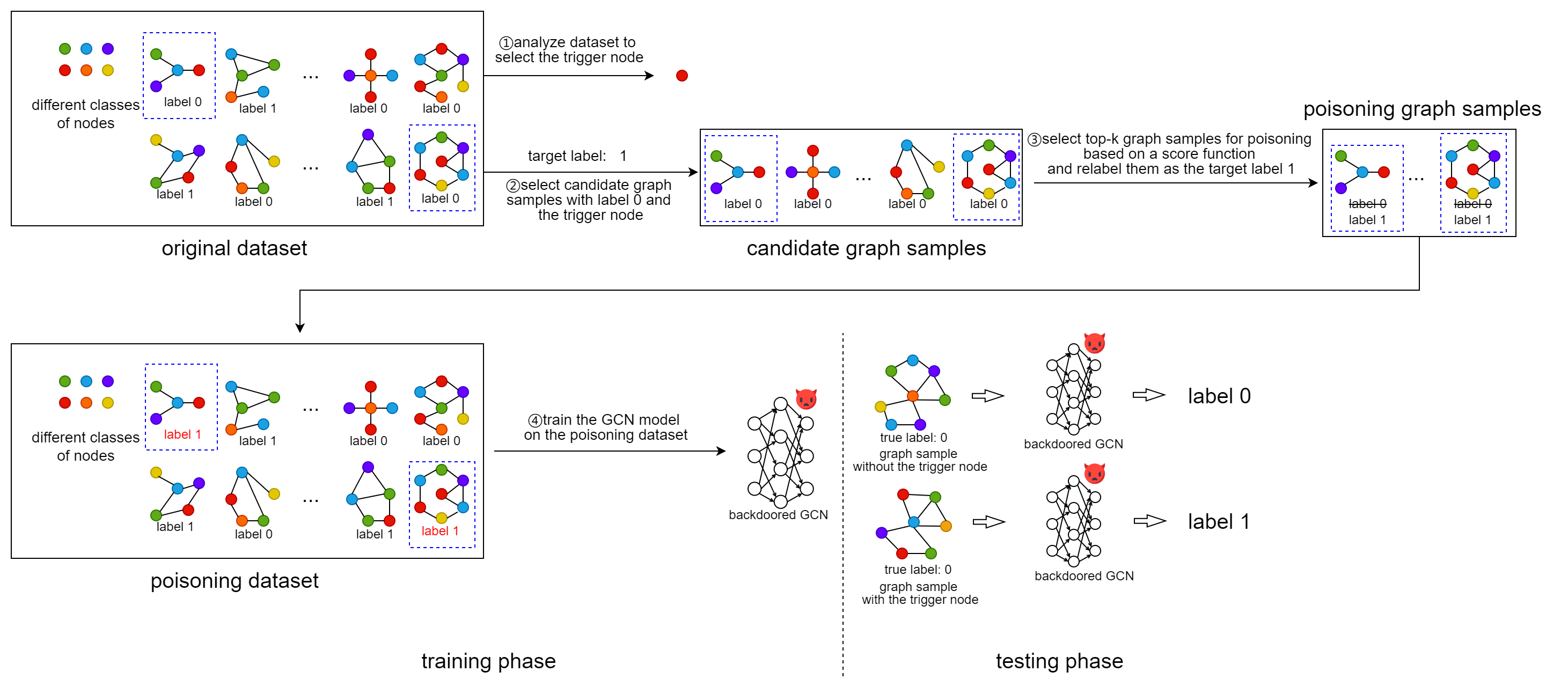}
    \caption{The implementation process of SBAG, which is illustrated by an example of binary classification. We assume there are six different classes of nodes in the samples of the original dataset, and they are represented in six different colors: green, blue, purple, red, orange and yellow. The labels of the samples are either 0 or 1, and the target label is 1. SBAG consists of four steps: 1. The attacker analyzes the samples in the original dataset to select the semantic trigger node. 2. The attacker generates the poisoning samples by selecting candidate graph samples with label 0 and the trigger node first. Then, the attacker further selects the samples with the top-k score based on a score function and relabels them as target label 1. 3. The training dataset becomes a poisoning dataset with poisoning samples, and this dataset is used to train the GCN model to embed the backdoor in the model. The two samples in the dashed box illustrate how poisoning samples are generated. 4. After the semantic backdoor is embedded into the GCN model, the input samples with enough trigger nodes will activate the backdoor in the model, and they will be predicted as the target label, while those without the trigger node will be predicted correctly.}
    \label{fig2}
\end{figure}

At a high level, the implementation process of SBAG is illustrated in Figure \ref{fig2}. For ease of explanation, the example in the figure is binary classification of graph samples, and the implementation process of the multiclassification scenarios is similar to this.

We assume there are six different classes of nodes in the samples of the original dataset, and they are represented in six different colors: green, blue, purple, red, orange and yellow. The labels of the samples are either 0 or 1, and the target label is 1. SBAG consists of the following four steps.

\begin{enumerate}
  \item \textbf{Selecting the semantic trigger node: }The attacker analyzes the samples in the original dataset to select the semantic trigger node, which will be described in detail in Section \ref{Selecting the semantic trigger node}. For example, the red node is selected as the trigger node in step one in the figure after analyzing the samples in the original dataset.
  \item \textbf{Generating the poisoning samples: }First, the attacker selects candidate samples from the original dataset that have labels other than the target label and contain the trigger node. Then, the attacker generates the poisoning samples by selecting samples with the top-k score based on a score function and relabeling them as the target label. For example, in the step two in the figure, the samples with label 0 and the red trigger nodes are selected as candidate samples from which the samples with top-k score based on a score function are further selected as the poisoning samples and their labels are changed to the target label 1, as shown in step three in the figure. The details of generating the poisoning samples will be presented in Section \ref{Poisoning samples generation}.
  \item \textbf{Training with poisoning data: }The poisoning samples are put into the training dataset to replace the original ones. For example, in Figure \ref{fig2}, two samples in the dashed box in the original dataset are selected as candidate samples first, then they are further selected as poisoning samples and relabeled from the original ground-truth label 0 to the target label 1. Finally, the training dataset becomes a poisoning dataset with these poisoning samples. The GCN model trained with the poisoning dataset embeds the semantic backdoor into the model, as shown in step four in the figure.
  \item \textbf{Activating the backdoor: }After the semantic backdoor is embedded into the GCN model, the input samples with enough trigger nodes will activate the backdoor in the model, and they will be predicted as the target label, while those without the trigger nodes will be predicted correctly. As shown in the testing phase in Figure \ref{fig1}, the upper sample with ground-truth label 0 but no trigger node will be classified correctly by the backdoored GCN model as label 0, while the lower samples with ground-truth label 0 and the trigger node will be misclassified as target label 1.
\end{enumerate}

The SBAG has the following assumptions:
\begin{enumerate}
    \item SBAG is aimed at the task of graph classification.
    \item The attacker can access the training set and modify the label of some samples.
    \item Every node in the graph samples has a class as its identifier. This assumption is without loss of generality. For example, nodes in a graph of protein molecules represent the basic elements such as carbon, hydrogen, oxygen, and nitrogen that make up proteins, so the labels of the nodes are identified as C(carbon), H(hydrogen), O(oxygen) and N(nitrogen), respectively; for a social network, nodes represent users, and the user’s ID is the identifier of each node.
\end{enumerate}

\subsection{Selecting the semantic trigger node}
\label{Selecting the semantic trigger node}
Since our attack uses a certain class of naturally occurring nodes in the graph samples as the trigger, we need to select the node as the trigger that has a strong association with the target label.

First, we count the number of each class of nodes in the training dataset. We denote $y_t$ as the target label and  $\bar{y_t}$ as nontarget labels of the samples. For each class of nodes $n_i$ in the dataset, we count how many graph samples with $y_t$ and $\bar{y_t}$ contain $n_i$, which are denoted as $num(n_i,y_t)$ and $num(n_i,\bar{y_t})$, respectively.

Then, we compute the number of poisoning samples according to the poisoning rate, which is denoted as ${num}_p$ and can be obtained as follows:
\begin{displaymath}
    {num}_p=|D| \ast p,
\end{displaymath}
where $|D|$ is the number of samples in the training set and $p$ is the poisoning rate.

Finally, we select the node $n_i$ as the trigger where $num(n_i,\bar{y_t})$ is the closest number to ${num}_p$ so that as many graph samples with the trigger and the label $\bar{y_t}$ as possible can be turned into poisoning samples by relabeling them to the target label $y_t$, which ensures that the trigger has a strong association with the target label $y_t$.

Takeing the AIDS dataset as an example, it is a binary dataset with 2000 graph samples, where 0 and 1 denote the labels of the samples ($y_t$ and $\bar{y_t}$, respectively). There are 38 classes of nodes in total which are indexed from 0 to 37, and the target label is 0. For each class of nodes $n_i(i=0,1\ldots,37)$ in the dataset, we count how many graph samples with $y_t$ and $\bar{y_t}$ contain $n_i$. The statistical table is as follows: for example, the numbers 400 and 1570 in the second row of the table represent that there are 400 graph samples with label 0 and node 0 $(n_0)$, and there are 1570 graph samples with label 1 and node 0 $(n_0)$.

\begin{table}[htbp]
\centering
\caption{An example on the AIDS dataset}
\label{An example on the AIDS dataset}
\begin{tabular}{|c|c|c|}
\hline
                         & $num(n_i,0)$  & $num(n_i,1)$  \\ \hline
node 0 ($n_0$)           & 400           & 1570          \\ \hline
node 1 ($n_1$)           & 385           & 1177          \\ \hline
...                      & ...           & ...           \\ \hline
\textbf{node 7 ($n_7$)}  & \textbf{42}   & \textbf{49}   \\ \hline
node 8 ($n_8$)           & 50            & 10            \\ \hline
...                      & ...           & ...           \\ \hline
node 37 ($n_{37}$)       & 0             & 1             \\ \hline
\end{tabular}
\end{table}

Assuming the poisoning rate $p$ is $2\%$, we can calculate the poisoning number ${num}_p=|D| \ast p=2000 \ast 2\%=40$ and choose node 7 as the trigger node $T$ because $num(n_7,\bar{y_t})=49$ is the closest number to ${num}_p=40$, as shown in bold font in the table. After that, we select 40 samples from the 49 graph samples that have the trigger node $T$ and label 1 and relabel them to the target label 0. How to select these graphs for poisoning will be described in the next section.

Algorithm \ref{alg1} sketches the flow of the above method.
\begin{algorithm}[htbp]
\caption{Finding the semantic trigger node}\label{alg1}
\begin{algorithmic}
\renewcommand{\algorithmicrequire}{\textbf{Input:}}
\renewcommand{\algorithmicensure}{\textbf{Output:}}
\Require $D$ - original training set; $y_t$ – target label; $p$ – poisoning rate;
\Ensure $T$ - the trigger node;
\ForAll{$g \in D$}
    \State nodeSet = set(); \Comment{Create an empty set to record all classes of nodes in $g$}
    \ForAll{$n_i \in g$}
        \State add $n_i$ to nodeSet;
    \EndFor
    \ForAll{$n_i \in nodeSet$}
        \State $num(n_i,g.label)=num(n_i,g.label)+1$;
    \EndFor
\EndFor
\State 	${num}_p = |D| \ast p$;  \Comment{compute the number of poisoning samples by $p$}
\ForAll{$i$}
    \State 	select $n_i$ as $T$ where $num(n_i,\bar{y_t})$ is the closest to ${num}_p$;
\EndFor
\State \textbf{return} $T$
\end{algorithmic}
\end{algorithm}

\subsection{Poisoning sample generation}
\label{Poisoning samples generation}
Once the trigger node is identified, the next step is to select graph samples from the training dataset to create poisoning samples.

First, we select candidate graph samples from the training dataset that have label $\bar{y_t}$ and semantic trigger node T and satisfy $num(T,\bar{y_t})$ being the closest to ${num}_p$. The candidate graph samples can be defined formally as follows:
\begin{displaymath}
    candidate\ samples=\{g\ |\ (g \in D)\ and\ (T \in g)\ and\ (g.label \neq y_t)\}
\end{displaymath}

Second, to select poisoning samples from the candidate graph samples, we train another GCN model called the scoring model $f_{score}$ for rating the candidate graph samples, and the top-k ($k={num}_p$) score candidate graph samples are selected as the poisoning samples. Specifically, we first train the scoring GCN model $f_{score}$ based on the original dataset, and the model's prediction results are the degree of confidence of the input sample on each class. To select candidate graph samples where the trigger nodes have a significant impact on their classification results. We modify the features of all trigger nodes in the graph samples to 0 without changing the graph topology. After that, the original and modified samples are predicted with $f_{score}$, and we can obtain the difference between their confidence on their original label. We take the absolute value of this difference as the score of the candidate graph sample. Formally, we define this operation as follows:
\begin{displaymath}
    sample\ score=|f_{score}(g)-f_{score}(g\prime)|,\ g \in candidate\ samples
\end{displaymath}
where g is the original graph sample in $candidate\ samples$, $g\prime$ is the modified graph sample after modifying features of trigger nodes, and $|\ \bullet\ |$ is an absolute value operation.

Finally, the top-k score candidate graph samples are selected as poisoning samples, and they are relabeled from $\bar{y_t}$ to the target label $y_t$.

\subsection{Training with poisoning data}
\label{Training with poisoning data}
The training dataset becomes a poisoning dataset with poisoning samples. After being well trained with the poisoned training dataset, the semantic backdoor will be embedded into the GCN model, which will associate the semantic backdoor trigger node with the target label $y_t$.

\subsection{Backdoor activation}
After the semantic backdoor is embedded into the GCN model, the model will misclassify the input sample as the target label $y_t$ as long as the input sample contains enough semantic trigger nodes. The backdoor in the model can be activated by original testing samples with the trigger nodes without any modification or by injecting a small number of the trigger nodes in the testing samples that do not contain the trigger nodes.
 
\section{Attack Evaluation}
\label{Attack Evaluation}
In this section, we evaluate SBAG on the graph classification task with three experiments. First, we test the classification accuracy of the clean GCN model and the backdoored GCN models with different poisoning rates on benign graph samples to evaluate the latter’s performance on benign samples. Second, we test the classification accuracy of the backdoored GCN model on original testing samples with the trigger to evaluate whether these samples can activate the backdoor in the model without any modification and the attack success rate of SBAG. Third, we test the classification accuracy of the backdoored GCN model on testing samples that originally do not contain any trigger nodes but are injected with trigger nodes to evaluate the attack success rate of SBAG. Finally, we compare these results with two state-of-the-art baselines to show that SBAG is effective.

\subsection{Experimental settings}
\textbf{Datasets - }We select three binary graph classification datasets and one multiclass graph classification dataset from TUDatasets, which is a collection of benchmark datasets for graph classification and regression \citep{morris2020tudataset}.

The binary graph classification datasets are the following: (i) AIDS \citep{riesen2008iam} – this dataset consists of 2000 graphs representing molecular compounds, and each graph is classified as “active” or “inactive”; (ii) NCI1 \citep{wale2008comparison} – this dataset comes from the cheminformatics domain, which is relative to anticancer screens where the chemicals are assessed as positive or negative to cell lung cancer; (iii) PROTEINS - a dataset of proteins that are classified as enzymes or nonenzymes \citep{borgwardt2005protein,dobson2003distinguishing};

The multiclass graph classification dataset is ENZYMES - a dataset of 600 protein tertiary structures obtained from the BRENDA enzyme database, consisting of 6 classes of enzymes \citep{schomburg2004brenda}. The dataset statistics are summarized in Table \ref{The dataset statistics}.

\begin{table}[htbp]
\centering
\caption{The dataset statistics}
\label{The dataset statistics}
\resizebox{\textwidth}{!}{
\begin{threeparttable}
\begin{tabular}{|l|l|l|l|l|l|l|}
\hline
Dataset & Graph num. & Avg. Nodes & Avg. edges & Class num. of graphs & Graph num. [Class] & Target class \\ \hline
AIDS     & 2000 & 15.69 & 16.20 & 2 & 400[0], 1600[1]  & 0 \\ \hline
NCI1     & 4110 & 29.87 & 32.30 & 2 & 2053[0], 2057[1] & 0 \\ \hline
PROTEINS & 1113 & 39.06 & 72.82 & 2 & 663[0], 450[1]   & 1 \\ \hline
ENZYMES  & 600  & 32.63 & 62.14 & 6 & 100 for each     & 5 \\ \hline
\end{tabular}
\begin{tablenotes}
\item[*] Graph num. - number of graphs in the dataset; Avg. Nodes - average number of nodes per graph; Avg. edges - average number of edges per graph; Class num. of graphs – the number of classes of graph samples; Graph num. [Class] – the number of graph samples in [class]
\end{tablenotes}
\end{threeparttable}
}
\end{table}

\textbf{Metrics - }We introduce the following three metrics to evaluate the effectiveness of SBAG.
\begin{enumerate}
    \item Attack success rate (ASR) refers to the percentage of samples with the trigger that are classified into the target label by the backdoored model. We test the ASR with a dataset containing only backdoor samples.
    \begin{displaymath}
        Attack\ Success\ Rate(ASR)=\frac{the\ number\ of\ attack\ samples\ predicted\ as\ the\ target\ label}{the\ total\ number\ of\ attack\ samples}
    \end{displaymath}
    \item 2.	Clean accuracy drop (CAD) refers to the difference between the classification accuracy of a backdoored model on benign samples and that of a clean model on benign samples. In the following text, we refer to the classification accuracy of the clean model on benign samples as \textit{“clean accuracy”} and the classification accuracy of the backdoored model on benign samples as \textit{“benign accuracy”}. The \textit{“benign accuracy”} should be close to \textit{“clean accuracy”} to hide the existence of the backdoor.
    \begin{displaymath}
        Clean\ Accuracy\ Drop(CAD)=clean\ accuracy-benign\ accuracy
    \end{displaymath}
    \item Poisoning rate (p), which is the ratio of the number of poisoning samples to the total number of the training set. The lower the poisoning rate is, the easier and stealthier the backdoor attack is.
\end{enumerate}

We evaluate SBAG with ASR and CAD under different poisoning rates.

\textbf{Baselines - }To evaluate our attack, we compare SBAG with two state-of-the-art baselines.
\begin{itemize}
    \item Graph trojaning attack (GTA): Xi et al. \citep{xi2021graph} proposed GTA, which is a backdoor attack that trojans GNNs and invokes malicious functions in downstream tasks using subgraphs as triggers tailored to individual graphs.
    \item Subgraph-based backdoor attack (Subgraph): Zhang et al. \citep{zhang2021backdoor} proposed a subgraph-based backdoor attack to GNNs for graph classification. When a predefined subgraph is injected into the testing graph, a GNN classifier will predict the target label specified by the attacker.
\end{itemize}

\textbf{Parameter settings - }We use a three-layer GCN model with one hidden layer as our target GCN model followed by a global mean pooling layer for graph-level feature aggregation and a softmax layer for graph classification. Table \ref{Parameter settings} shows the parameter settings for our experiments. For GTA and subgraph, we use the same parameters as those provided in their original papers \citep{xi2021graph,zhang2021backdoor}.

\begin{table}[htbp]
\centering
\caption{Parameter settings}
\label{Parameter settings}
\begin{tabular}{|l|l|}
\hline
Parameter       & Settings                            \\ \hline
Architecture    & three-layer GCN (one hidden layer)  \\
Hidden channels & 32                                  \\
Pooling layer   & global\_mean\_pool                  \\
Classifier      & Softmax                             \\
Optimizer       & Adam                                \\
Weight decay    & 5e-4                                \\
Learning rate   & 0.01                                \\
Batch size      & 32                                  \\
Max epoch       & 100                                 \\ \hline
\end{tabular}
\end{table}

\subsection{Experimental Method}
For each dataset, we conduct our experiments through the following steps:
\begin{enumerate}
    \item \textbf{Test the classification accuracy of the clean model on benign samples: }We use stratified sampling by graph labels to randomly split the original dataset into two parts: 80\% of the dataset is used to train the clean model, and the remaining 20\% is used to test the classification accuracy of the clean model on benign samples.
    \item \textbf{Construct poisoning training sets under different poisoning rates and train backdoored models: }we randomly split the original dataset in an 80:20 ratio, and 80\% of the original dataset is used to construct the poisoning training dataset under different poisoning rates to train the backdoored models, as described in Section \ref{Poisoning samples generation} and Section \ref{Training with poisoning data}.The remaining 20\% of the original dataset is used to test the benign accuracy and the ASR of SBAG.
    \item \textbf{Test the classification accuracy of backdoored models on benign samples and unmodified original testing samples with the trigger nodes: }As mentioned above, we use 20\% of the original dataset to test the benign accuracy and the ASR on original testing samples with the trigger. Specifically, the graph samples from the 20\% of the original dataset that have label $\bar{y_t}$ and the trigger nodes are used to test ASR on original testing samples with the trigger, and the remaining samples of the 20\% of the original dataset are used to test benign accuracy on benign samples. The CADs are further calculated based on the benign accuracy and the clean accuracy obtained in the first step.
    \item \textbf{Test the classification accuracy of the backdoored models on original testing samples modified to inject triggers: }We select some samples from 20\% of the original dataset that have label $\bar{y_t}$ but do not have trigger nodes and inject an average number of trigger nodes in random places without modifying the topology of the samples to test the ASR of the backdoored models on original testing samples modified to inject triggers.
    \item \textbf{Comparison with baselines: }we compare all ASRs and CADs mentioned above with baselines to show that SBAG is effective.
\end{enumerate}

\subsection{Experimental Results}
\begin{enumerate}
    \item \textbf{The classification accuracy of the clean model on benign samples:}

    Table \ref{The Classification Accuracy of the Clean GCN Model on Benign Samples} shows the classification accuracy of the clean model on benign samples. Each classification accuracy is obtained by averaging the results of three repeated runs. For each dataset, we use stratified sampling by graph labels to randomly split the original dataset in an 80:20 ratio. The samples from 80\% of the original dataset are used to train the clean model, and the remaining 20\% of the samples of the original dataset are used to test the classification accuracy of the clean model on the benign samples.
    \begin{table}[htbp]
    \centering
    \caption{The Classification Accuracy of the Clean GCN Model on Benign Samples}
    \label{The Classification Accuracy of the Clean GCN Model on Benign Samples}
    \begin{tabular}{|c|c|}
    \hline
    Dataset  & Average classification accuracy (\%) \\ \hline
    AIDS     & 98.83                                \\ \hline
    NCI1     & 67.52                                \\ \hline
    PROTEINS & 70.70                                \\ \hline
    ENZYMES  & 40.28                                \\ \hline
    \end{tabular}
    \end{table}

    \item \textbf{The classification accuracy of the backdoored models on unmodified original testing samples with the trigger nodes under different poisoning rates:}
    
    The results are shown in Table \ref{The clearn accuracy, the benign accuracy and ASR of the unmodified original testing samples with the trigger nodes under different poisoning rates}. The results under the highest poisoning rates are highlighted in bold. From the table, we can see that the ASRs reach 99.9\% for three datasets: AIDS, NCI1 and PROTEINS when the poisoning rate is 3\%, and CADs are less than 1\%. For the ENZYMES dataset, a poison rate of 5\% is required to reach the maximum ASR with a 7.37\% drop in CAD.

    Figure \ref{fig3} is the visualization of Table \ref{The clearn accuracy, the benign accuracy and ASR of the unmodified original testing samples with the trigger nodes under different poisoning rates}. The horizontal axis of each graph represents the poisoning rate, and the vertical axis represents the percentage of ASR,clean accuracy and benign accuracy. In each graph, the ASR, clean accuracy and benign accuracy are represented by a red line, a blue line and a green line, respectively, and CAD can be indicated by the gap of the coordinates of the blue line and green line under the same poisoning rate. From the figure, we can see that on all four datasets, the ASRs continuously increase with increasing poisoning rate and reach their maximum values at poisoning rates of 3\% and 5\%. We can also see that the gap between blue lines and green lines is not significant in most positions except the ENZYMES dataset with a poisoning rate of 5\%.
    
    From the above results, we can obtain the following three conclusions: (1) The original testing samples with trigger nodes can activate the backdoor in the backdoored GCN model without modification; (2) SBAG achieves an attack success rate of approximately 99.9\% with a poisoning rate of less than 5\%; and (3) the backdoored GCN models have prediction accuracy close to that of the clean model on benign samples.

    \begin{table}[htbp]
    \centering
    \caption{The clean accuracy, benign accuracy and ASR of the unmodified original testing samples with the trigger nodes under different poisoning rates}
    \label{The clearn accuracy, the benign accuracy and ASR of the unmodified original testing samples with the trigger nodes under different poisoning rates}
    \begin{threeparttable}
    \begin{tabular}{|c|c|c|c|c|c|}
    \hline
    Dataset                   & Clean accuracy (\%)    & p               & Benign accuracy. (\%) & ASR (\%)      & CAD (\%)      \\ \hline
    \multirow{3}{*}{AIDS}     & \multirow{3}{*}{98.83} & 1.00\%          & 97.95                 & 20.77         & 0.88          \\ \cline{3-6} 
                              &                        & 2.00\%          & 98.46                 & 93.64         & 0.37          \\ \cline{3-6} 
                              &                        & \textbf{3.00\%} & \textbf{98.23}        & \textbf{99.9} & \textbf{0.6}  \\ \hline
    \multirow{3}{*}{NCI1}     & \multirow{3}{*}{67.52} & 1.00\%          & 68.03                 & 78.64         & 0             \\ \cline{3-6} 
                              &                        & 2.00\%          & 66.81                 & 98.48         & 0.71          \\ \cline{3-6} 
                              &                        & \textbf{3.00\%} & \textbf{70.96}        & \textbf{99.9} & \textbf{0}    \\ \hline
    \multirow{3}{*}{PROTEINS} & \multirow{3}{*}{70.70} & 1.00\%          & 71.73                 & 43.89         & 0             \\ \cline{3-6} 
                              &                        & 2.00\%          & 69.05                 & 66.18         & 1.65          \\ \cline{3-6} 
                              &                        & \textbf{3.00\%} & \textbf{70.94}        & \textbf{99.9} & \textbf{0}    \\ \hline
    \multirow{4}{*}{ENZYMES}  & \multirow{4}{*}{40.28} & 1.00\%          & 38.10                 & 32.38         & 2.18          \\ \cline{3-6} 
                              &                        & 2.00\%          & 38.91                 & 43.65         & 1.37          \\ \cline{3-6} 
                              &                        & 3.00\%          & 41.47                 & 61.11         & 0             \\ \cline{3-6} 
                              &                        & \textbf{5.00\%} & \textbf{32.91}        & \textbf{99.9} & \textbf{7.37} \\ \hline
    \end{tabular}
    \begin{tablenotes}
    \item[*] Clean accuracy - the classification accuracy of the clean model on benign samples; p - poisoning rate; Benign accuracy - classification accuracy of a backdoored model on benign samples
    \end{tablenotes}
    \end{threeparttable}
    \end{table}

    \begin{figure}[htbp]
    \centering
    \includegraphics[width=\textwidth]{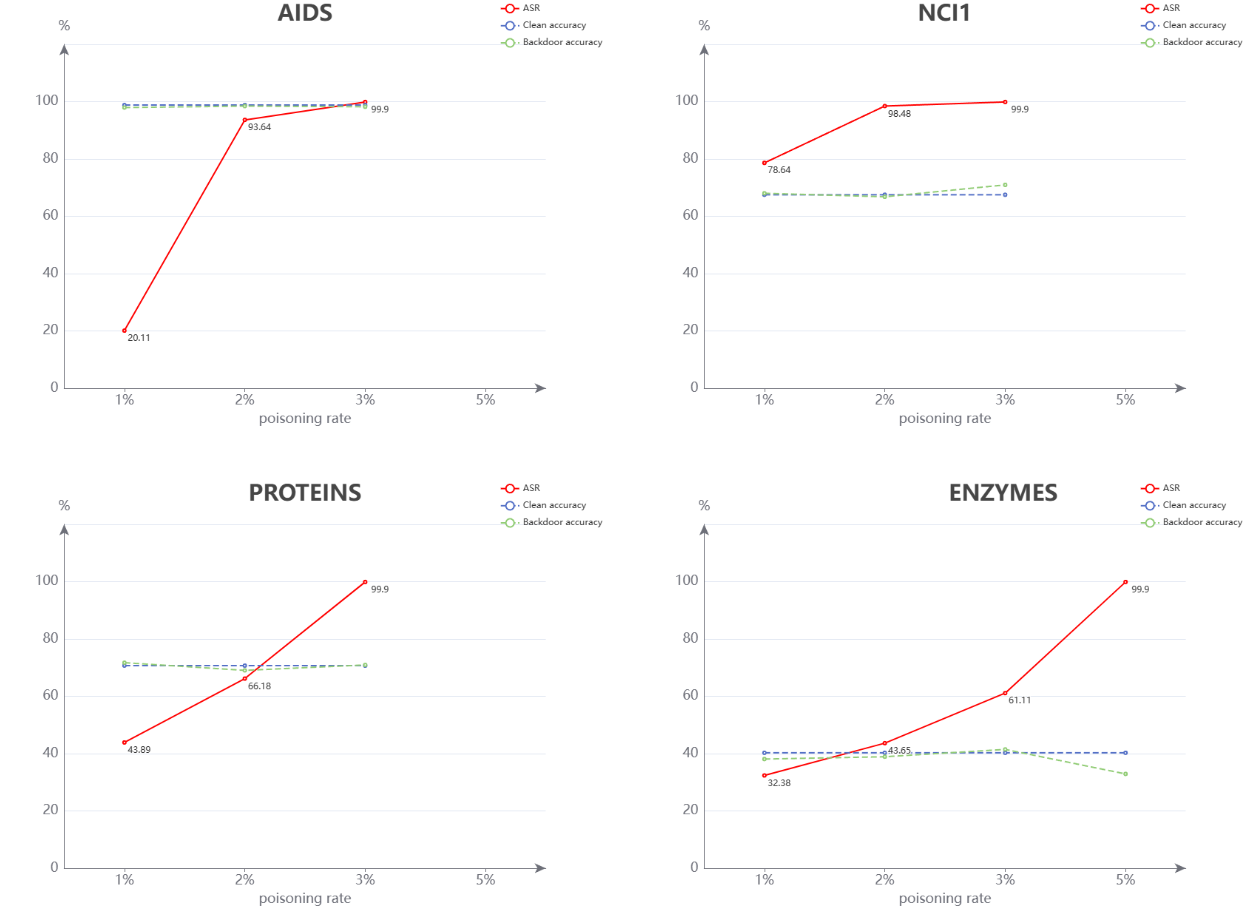}
    \caption{Impact of poisoning rate on ASR and CAD on four datasets.}
    \label{fig3}
    \end{figure}

    \item \textbf{The classification accuracy of the backdoored model on graph samples modified to inject trigger:}

    For those graph samples that have label $\bar{y_t}$ and do not contain any trigger nodes, we investigate whether such graph samples can activate the backdoor in the infected GCN model by injecting $k$ trigger nodes in each of them, where $k$ refers to the average number of trigger nodes appearing in each original graph sample and is defined as follows:
    \begin{displaymath}
        k=\frac{the\ sum\ of\ the\ number\ of\ T\ in\ each\ graph\ sample\ }{number\ of\ graphs\ containing\ T}.
    \end{displaymath}
    We randomly replace the features of $[k]$ and $[k]+1$ node(s) ($[\ ]$ represents the round down symbol) in each of the graph samples with those of the trigger node without modification of their topology and then predict it by $f_\theta^\prime$. If the prediction result is $y_t$, it indicates that the backdoor in the model is activated. Because the trigger node may have various feature vectors, we perform the above operations with each of the feature vectors and average the results.

    Table \ref{ASRs of graph samples modified to inject the trigger} shows the ASR of the infected model trained with the highest poisoning rate in our experiment on each dataset. From the table, we can see that by randomly modifying $[k]+1$ nodes of the samples to the trigger node, the backdoor in the infected model will be activated, and the model will misclassify these samples to the target label with high probabilities. Specifically, the ASRs on the AIDS and NCI1 datasets can achieve nearly 99.9\%, and the ASRs are lower on the PROTEINS and ENZYMES datasets due to their more complex graph structures, reaching 82.02\% and 87.96\%, respectively.

    As we just inject trigger nodes randomly in the samples without modification of the topology, SBAG achieves high attack success rates. We can conclude from the results that SBAG is effective on graph samples modified to inject triggers.

    \begin{table}[htbp]
    \centering
    \caption{ASRs of graph samples modified to inject the trigger}
    \label{ASRs of graph samples modified to inject the trigger}
    \begin{threeparttable}
    \begin{tabular}{|c|c|c|c|c|}
    \hline
    Dataset                   & p                    & k                      & Num. of modified nodes & ASR (\%) \\ \hline
    \multirow{2}{*}{AIDS}     & \multirow{2}{*}{3\%} & \multirow{2}{*}{1.37}  & 1                      & 97.96    \\ \cline{4-5} 
                              &                      &                        & 2                      & 99.87    \\ \hline
    \multirow{2}{*}{NCI1}     & \multirow{2}{*}{3\%} & \multirow{2}{*}{2.72}  & 2                      & 99.88    \\ \cline{4-5} 
                              &                      &                        & 3                      & 99.9     \\ \hline
    \multirow{2}{*}{PROTEINS} & \multirow{2}{*}{3\%} & \multirow{2}{*}{15.78} & 15                     & 81.07    \\ \cline{4-5} 
                              &                      &                        & 16                     & 82.02    \\ \hline
    \multirow{2}{*}{ENZYMES}  & \multirow{2}{*}{5\%} & \multirow{2}{*}{10.65} & 10                     & 86.78    \\ \cline{4-5} 
                              &                      &                        & 11                     & 87.96    \\ \hline
    \end{tabular}
    \begin{tablenotes}
    \item[*] p - poisoning rate; Num. of modified nodes - the number of nodes actually modified; the numbers are [k] and [k]+1.
    \end{tablenotes}
    \end{threeparttable}
    \end{table}

    \item \textbf{Comparison with baselines:}

    Table \ref{Comparison of SBAG with two baselines} summarizes the performance of SBAG and two baselines (GTA and subgraph) under the same poisoning rate. The highest ASR and the lowest CAD on each dataset are highlighted with shadow.
    
    From the table, we can see that SBAG can achieve higher ASRs than Subgraph on four datasets, whether on unmodified samples with the trigger nodes or samples modified to inject trigger nodes. In comparison with GTA, the ASRs of SBAG on unmodified samples with trigger nodes achieve the same results as those of GTA, reaching 99.9\% on four datasets. The ASRs of SBAG on samples modified to inject trigger nodes from datasets AIDS and NCI1 are nearly close to that of GTA (99.9\%). Although there is a slight decrease in the ASRs of SBAG on samples modified to inject trigger nodes from datasets PROTEINS and ENZYMES, they are over 82\%, reaching 82.02\% and 87.96\%, respectively. The reason for this may be that the graph structures in PROTEINS and ENZYMES are more complex, and we modify the features of nodes randomly selected to that of the trigger nodes without modifying the topology of these samples, which may have an impact on the backdoor attacks.
    
    The CADs of SBAG are better than those of GTA on the NCI1, PROTEINS and ENZYMES datasets, except for a slight decline on the AIDS dataset. Although Subgraph achieves better CADs than SBAG on AIDS and ENZYMES, the largest CAD of SBAG is less than 8\% on ENZYMES and there is almost no decline on the CADs of SBAG on other three datasets AIDS, NCI1, PROTEINS (0.8\%, 0\% and 0\%).
    
    From the above comparison, we can see that SBAG is effective with its high ASRs and low CADs, which means that SBAG, as a black-box semantic backdoor attack, is easier to implement in reality and more imperceptible.

    \begin{table}[htbp]
    \centering
    \caption{Comparison of SBAG with two baselines}
    \label{Comparison of SBAG with two baselines}
    \resizebox{\textwidth}{!}{
    \begin{threeparttable}
    \begin{tabular}{|c|c|ccc|cc|cc|}
    \hline
     &
       &
      \multicolumn{3}{c|}{SBAG (ours)} &
      \multicolumn{2}{c|}{GTA} &
      \multicolumn{2}{c|}{Subgraph} \\ \cline{3-9} 
     &
       &
      \multicolumn{2}{c|}{ASR (\%)} &
       &
      \multicolumn{1}{c|}{} &
       &
      \multicolumn{1}{c|}{} &
       \\ \cline{3-4}
    \multirow{-3}{*}{Dataset} &
      \multirow{-3}{*}{p} &
      \multicolumn{1}{c|}{unmodified} &
      \multicolumn{1}{c|}{modified} &
      \multirow{-2}{*}{CAD (\%)} &
      \multicolumn{1}{c|}{\multirow{-2}{*}{ASR (\%)}} &
      \multirow{-2}{*}{CAD (\%)} &
      \multicolumn{1}{c|}{\multirow{-2}{*}{ASR (\%)}} &
      \multirow{-2}{*}{CAD (\%)} \\ \hline
    AIDS &
      3\% &
      \multicolumn{1}{c|}{\cellcolor[HTML]{D0CECE}99.9} &
      \multicolumn{1}{c|}{\cellcolor[HTML]{FFFFFF}99.87} &
      0.8 &
      \multicolumn{1}{c|}{\cellcolor[HTML]{D0CECE}99.9} &
      \cellcolor[HTML]{D0CECE}0 &
      \multicolumn{1}{c|}{92.12} &
      0.3 \\ \hline
    NCI1 &
      3\% &
      \multicolumn{1}{c|}{\cellcolor[HTML]{D0CECE}99.9} &
      \multicolumn{1}{c|}{\cellcolor[HTML]{D0CECE}99.9} &
      \cellcolor[HTML]{D0CECE}0 &
      \multicolumn{1}{c|}{\cellcolor[HTML]{D0CECE}99.9} &
      0.98 &
      \multicolumn{1}{c|}{99.85} &
      2.7 \\ \hline
    PROTEINS &
      3\% &
      \multicolumn{1}{c|}{\cellcolor[HTML]{D0CECE}99.9} &
      \multicolumn{1}{c|}{\cellcolor[HTML]{FFFFFF}82.02} &
      \cellcolor[HTML]{D0CECE}0 &
      \multicolumn{1}{c|}{\cellcolor[HTML]{D0CECE}99.9} &
      6.5 &
      \multicolumn{1}{c|}{30.91} &
      3.13 \\ \hline
    ENZYMES &
      5\% &
      \multicolumn{1}{c|}{\cellcolor[HTML]{D0CECE}99.9} &
      \multicolumn{1}{c|}{\cellcolor[HTML]{FFFFFF}87.96} &
      7.37 &
      \multicolumn{1}{c|}{\cellcolor[HTML]{D0CECE}99.9} &
      10.56 &
      \multicolumn{1}{c|}{24.55} &
      \cellcolor[HTML]{D0CECE}1.5 \\ \hline
    \end{tabular}
    \begin{tablenotes}
    \item[*] p - poisoning rate
    \end{tablenotes}
    \end{threeparttable}
    }
    \end{table}
    
\end{enumerate}

\section{Conclusion}
\label{Conclusion}
In this work, we propose a black-box semantic backdoor attack called SBAG against graph convolutional networks, which reveals that GCNs are vulnerable to the attacks. SBAG uses naturally occurring nodes as triggers whichthat exist in the original dataset. When trigger nodes appear in the graph sample, the backdoored model will misclassify it to the target label specified by the attacker. Our experimental evaluation results on four real-world datasets show that samples with the semantic trigger can activate the backdoor in the infected model with high probability even without the requirement of modification and can achieve a high attack success rate with a poisoning rate of less than 5\%. In the future, we will investigate how gradients change in the backdoored graph convolutional networks and Study study a defense method against our semantic backdoor attack.

\section{Acknowledgement}
Acknowledgement:the research of the paper was supported by Natural Science Foundation of Shanghai Municipality (Grant NO.22ZR1422600)

\bibliographystyle{elsarticle-num}
\bibliography{main.bib}

\end{document}